# Identifying Patterns of Associated-Conditions through Topic Models of Electronic Medical Records

Moumita Bhattacharya[1], Claudine Jurkovitz, MD, MPH[2] and Hagit Shatkay, PhD[1,3,4]

[1]Computational Biomedicine Lab, Computer and Information Sciences, University of Delaware, Newark, DE, USA

[2]Value Institute, Christiana Care Health System, Newark, DE, USA

[3]Center for Bioinformatics and Computational Biology, Delaware Biotechnology Inst, University of Delaware, Newark, DE, USA

[4]School of Computing, Queen's University, Kingston, ON, K7L 3N6, Canada

{moumitab, shatkay}@udel.edu

*Abstract*— Multiple adverse health conditions co-occurring in a patient are typically associated with poor prognosis and increased office or hospital visits. Developing methods to identify patterns of co-occurring conditions can assist in diagnosis. Thus, identifying patterns of association among co-occurring conditions is of growing interest. In this paper, we report preliminary results from a data-driven study, in which we apply a machine learning method, namely, *topic modeling*, to *Electronic Medical Records* (*EMRs*), aiming to identify patterns of associated conditions. Specifically, we use the well-established *Latent Dirichlet Allocation* (*LDA*), a method based on the idea that documents can be modeled as a mixture of latent topics, where each topic is a distribution over words. In our study, we adapt the LDA model to identify latent topics in patients' EMRs. We evaluate the performance of our method both qualitatively and quantitatively, and show that the obtained topics indeed align well with distinct medical phenomena characterized by co-occurring conditions.

## I. INTRODUCTION

According to the Centers for Disease Control and Prevention, one in four individuals in the United States suffers multiple health conditions, while the rate is even higher (three in four), among individuals who are 65 or older. Per capita healthcare expenditure increases sharply as the number of conditions increases [1]. Patients suffering from multiple conditions pose a challenge to healthcare service providers as their prognosis is often poor and their visits frequency to primary care providers and hospitals is increased. Thus, identifying co-occurrence patterns of medical conditions is of growing interest, as it can help build accurate prediction models for hospitalization, progression of disease, or death. We report here preliminary results from a data-driven study in which we apply topic modeling to *Electronic Medical Records* (*EMRs*), aiming to identify patterns of associated conditions.

We conduct our analysis on a dataset comprising EMRs of patients obtained from multiple primary care practices in the State of Delaware. A total of 13,111 patient records were included in this study. They represent patients whose kidney function is decreased, as indicated by lower than normal (below 60 mL/min/m2) estimated *Glomerular filtration rate* (*GFR*), which is a common marker of kidney function. Each record includes attributes such as *age*, *gender*, *lab test results* and *diagnosed conditions*, recorded during multiple visits over a period of eight years.

We focus our analysis only on the *diagnosed conditions* attribute in the EMR dataset, represented through the healthcare terminology of *SNOMED-CT codes* [2], a common standardized language to record diagnosed conditions in EMRs, across different healthcare providers. SNOMED-CT is specifically designed to capture detailed information during clinical care by enabling clinicians to choose appropriate conditions from a predefined fine-grained list. The large number of patients, the wide timespan in our EMRs and the use of SNOMED codes to represent diagnosed conditions give rise to a large-scale dataset suitable for identifying patterns of co-occurring conditions.

Topic modeling is primarily used for identifying latent topics in a set of documents, based on the idea that documents can be modeled as a mixture of latent topics, where each topic is a distribution over words. In our study, a patient file, comprising all coded conditions with which the patient has been diagnosed, is viewed as a document, and each code is treated as a word. We use a well-established topic modeling technique, *Latent Dirichlet Allocation* (*LDA*) [3] to model patient files as though they were generated as a mixture of $K$ underlying topics, where a topic is a probability distribution over SNOMED codes; each code is assigned a probability to be associated with each topic. We hypothesize that the coded conditions that show a high probability to be associated with a specific topic, indeed tend to co-occur in patients.

Previous studies in other domains have employed topic models for a variety of natural language processing and image processing applications [3, 4]. Recently, topic models have also been applied in the biomedical domain for case-based retrieval [5], characterization of clinical concepts over time [6], and prediction of patient satisfaction and mortality [7, 8], among others. Topic models have also been employed to analyze differences in language use between depressed and non-depressed individuals [9], as well as to rank gene-drug relationships in the biomedical literature [10]. The majority of previous applications have centered around text data. To our knowledge, only a handful of studies have applied topic models to non-text data [11, 12]. However, compared to these studies, we analyze a much larger dataset and take a more rigorous approach to assess the clinical relevance of our results and to quantitatively evaluate the performance of our method.

We evaluate the performance of our method using two approaches. First, we assess the medical validity of our results by examining whether the conditions that show a high probability to be associated with a topic are known to co-occur according to the medical literature. Second, we quantitatively assess the topics obtained from our model by measuring how distinct they are from one another (*distinctiveness*) and whether a topic can be specified by a small number of conditions (*tightness*). We measure *distinctiveness* by calculating inter-topic distance using *Jensen-Shannon divergence* [13] – a symmetric measure of similarity between two probability distributions. *Tightness* is measured by inspecting, for each topic, the number of associated codes whose probability is greater than a threshold value; a low number of associated codes indicates that a topic can be characterized by a handful of codes, and is thus tight. Our results show that the topics are indeed *distinct* and *tight*, while aligning well with sets of conditions that are known to co-occur according to the medical literature.

The rest of the paper is organized as follows: Section II describes the experimental setting, including the dataset used, the data preprocessing steps, the LDA algorithm and the Jensen-Shannon divergence measure; Section III presents and discusses the results; Section IV summarizes our findings and proposes directions for future work.

## II. EXPERIMENTAL SETTING

Our dataset consists of information gathered from EMRs of patients in the state of Delaware, showing evidence of decrease in kidney function, recorded during office visits to physicians. Specifically, patients were included in the dataset if at least one estimated GFR value in their records was below *60 mL/min/m2*. The EMRs contain several attributes such as *age*, *ethnicity*, *gender*, *lab test results* and *diagnosed conditions* collected between August 2007 and July 2015 for 13,111 patients. The average number of visits per patient, over the 8-year period, is 6. The age range of patients in our dataset is 18-107, where 70% of the patients are 58-82 and the mean age is 70 ($\sigma$ =12.4). Our dataset consists of 60% female and 40% male patients. In this study, we focus solely on the diagnosed conditions attribute, represented via SNOMED codes. Table I lists the ten most frequent SNOMED codes in our dataset, along with their description and occurrence frequency.

We preprocess and organize the data to form records that can be used to fit a topic model. Typically in topic modeling, a word is the basic unit of discrete data while the set of unique words is referred to as the *vocabulary*. In contrast, in our study, we use *diagnosed conditions*, represented as SNOMED codes, rather than words, such that the set of unique codes forms our vocabulary. To determine the set of codes included, we create a list in which each code is associated with the number of times it occurs within the dataset, and note that 180 of the 5,000 codes present in the dataset account for 80% of the cumulative frequency. To avoid sparsity in the dataset, we limit our vocabulary to these 180 most frequent SNOMED codes.

Based on this vocabulary, we create a data matrix where rows correspond to patient-IDs and columns correspond to SNOMED codes, such that each cell <*p*, *c*> in the matrix contains the number of times a patient *p* was diagnosed with condition *c*. Thus, each patient is associated with a 180-dimensional vector, in which each entry represents the occurrence frequency of a diagnosed condition within the patient's record. We refer to each such vector as a *patient-conditions record* and to the collection of all such vectors as the *patient-conditions corpus*, represented by a matrix of dimension 13,111 by 180.

TABLE I. THE TEN MOST FREQUENT SNOMED CODES IN OUR EMR DATASET

| SNOMED Code | SNOMED Description | Number of Occurrences |
|---|---|---|
| 1201005 | Benign Essential Hypertension | 148,424 |
| 55822004 | Hyperlipidemia | 82,890 |
| 44054006 | Type 2 Diabetes Mellitus | 59,156 |
| 235595009 | Gastroesophageal Reflux | 48,731 |
| 267432004 | Pure Hypercholesterolemia | 47,022 |
| 414916001 | Obesity | 40,499 |
| 61582004 | Allergic Rhinitis | 40,066 |
| 40930008 | Hypothyroidism | 39,534 |
| 53741008 | Coronary Arteriosclerosis | 36,795 |
| 271795006 | Malaise and Fatigue | 27,581 |

***Latent Dirichlet Allocation***: LDA is a generative probabilistic model based on the idea that documents can be modeled as a mixture over latent topics, where each topic is a distribution over words [3]. We employ LDA to model patient records as though they were generated by sampling from a mixture of *K* underlying topics, where a topic is a multinomial distribution over all SNOMED codes in our vocabulary. Each code is thus viewed as a sample from a multinomial distribution over codes, and each such multinomial is selected from a distribution over *K* topics. By inferring the probability distributions associated with the *K* topics, we can characterize patient records as multinomial distributions over codes.

The number of patients in our corpus is *13,111*. The number of unique codes that form our vocabulary is denoted by *V*; in the experiments reported here *V=180*. We represent a patient's record as a *V*-dimensional vector of SNOMED codes, referred to as a *patient-conditions record*. The patient-conditions records are obtained by preprocessing the original patient file in the EMR; a patient file comprises all coded conditions with which the patient had been diagnosed during the 8-years period reflected in our dataset.

We represent each patient file, $F_i$ (*1 ≤ i ≤ 13,111*), as a vector of codes $F_i = (c_1^i, ..., c_{N_i}^i)$, where $N_i$ is the total number of code occurrences in the $i^{th}$ patient file. Each code, $c_j^i$, in the vector is one of the *V* SNOMED codes in our vocabulary, viewed as a value taken by a respective random variable, $C_j$ (*1≤ j ≤ N_i*), denoting the code value occurring in the $j^{th}$ position of the $i^{th}$ patient file. We note that any of the *V* codes in our vocabulary can appear at any position in a patient file. The generative process for each patient file consists of the following steps: First, a multinomial distribution over *V* codes for the $t^{th}$ topic, denoted $\Phi_t$ ($1 \le t \le K$), is obtained by sampling from a Dirichlet distribution with parameter $\alpha$; $\Phi_t$ represents the conditional probability of a code to occur in the $t^{th}$ topic. Next, for each patient file, $F_i$, a multinomial distribution over $K$ topics, denoted $\theta_i$, is sampled from a Dirichlet distribution with parameter $\beta$; $\theta_i$ represents the conditional probability of the file to be associated with each of the $K$ topics. Subsequently, for each code-position, $j$, in the file, $F_i$ : (*1*) A topic is drawn by sampling from $\theta_i$; the selected topic at position $j$ in the file $F_i$ is denoted $z_j^i \in \{1,...,K\}$; (*2*) Given the topic $z_j^i$ a code, $c_j^i$, is drawn by sampling from the topic-code distribution, $\Phi_{z_j^i}$.

To learn the model parameters based on our data, we use the R *topicmodels* library [14].

***Jensen-Shannon Divergence***: The *Jensen-Shannon divergence* (*JSD*) [13] is a symmetric measure of similarity between two probability distributions. Let $\vec{X} = <x_1,..., x_N>$ and $\vec{Y} = <y_1,..., y_N>$ be two *N*-dimensional vectors that represent two discrete probability distributions. The Jensen-Shannon divergence between $\vec{X}$ and $\vec{Y}$ is defined as:
$JSD(\vec{X}||\vec{Y}) = \frac{1}{2}\sum_{i=1}^{N} x_i \log\left(\frac{x_i}{m_i}\right) + \frac{1}{2}\sum_{i=1}^{N} y_i \log\left(\frac{y_i}{m_i}\right)$, where the vector $\vec{m} = <m_1,..., m_N>$ is a *N*-dimensional vector representing the mean distribution of $\vec{X}$ and $\vec{Y}$, calculated as: $m_i = \frac{1}{2}(x_i + y_i)$.

The JSD values range between 0 and *ln(2)* (~0.693), where 0 indicates identical distributions, and *ln(2)* indicates non-overlapping distributions. We use the JSD to calculate the inter-topic distance between each pair of topics, where the distribution dimension *N* is 180 (the number of codes in our vocabulary).

### III. EXPERIMENTS AND RESULTS

***Experiments:*** We applied LDA to the patient-conditions corpus to obtain topics, where each topic is a distribution over SNOMED codes. We ran multiple experiments varying the number of topics, and focus here on results obtained when using 20 topics. To ensure an appropriate *burn-in period*, which is the initial stage of the sampling process when the Gibbs samples are poor estimates of the posterior, we discarded the first 4,000 samples, after which we saved 4,000 Gibbs samples at regular intervals of 100 [15]. We used the default values, set in the *topicmodels* library, for the parameter $\beta$ (0.1) and for the initial value of the parameter $\alpha$ (50/*M*) [14].

Table II shows examples of six characteristic topics from the 20 identified by our model. For each of the six topics, we list the ten diagnosed conditions that have the highest probability to occur in the topic, along with their respective probabilities. We display only ten diagnosed conditions, since for most topics, the cummulative probability mass associated with these conditions accounts for over 0.9 of the total probability mass, as shown at the bottom row of the table. Moreover, the remaining diagnosed conditions have probability lower than 0.01. Similar results were obtained for the other 14 topics.

***Medical Relevance:*** To evaluate whether topic modeling indeed identifies patterns of association among patients' conditions, we verify that the most probable conditions within each topic are indeed known to co-occur according to the medical literature.

TABLE II. EXAMPLES OF SIX CHARACTERISTIC TOPICS FROM THE TWENTY IDENTIFIED BY OUR MODEL; EACH COLUMN LISTS TEN DIAGNOSED CONDITIONS THAT HAVE THE HIGHEST PROBABILITIES TO BE ASSOCIATED WITH THE RESPECTIVE TOPIC, ALONG WITH THEIR PROBABILITIES

| Topic A | Prob | Topic B | Prob | Topic C | Prob | Topic D | Prob | Topic E | Prob | Topic F | Prob |
|---|---|---|---|---|---|---|---|---|---|---|---|
| Type 2 diabetes mellitus | .369 | Pain in limb | .195 | Asthma | .200 | Allergic rhinitis | .361 | Hypothyroidism | .400 | Obesity | .383 |
| Type II diabetes mellitus uncontrolled | .132 | Arthralgia of the lower leg | .166 | Cough | .164 | Osteoporosis | .186 | Disorder of bone and articular cartilage | .229 | Depressive disorder | .250 |
| Mixed hyperlipidemia | .099 | Low back pain | .144 | Benign neoplasm of colon | .145 | Acute sinusitis | .111 | Vitamin D deficiency | .161 | Obstructive sleep apnea syndrome | .119 |
| Disorder associated with type 2 diabetes mellitus | .088 | Shoulder joint pain | .128 | Acute bronchitis | .117 | Benign essential hypertension | .110 | Anemia | .078 | Benign essential hypertension | .092 |
| Neurologic disorder associated with type 2 diabetes mellitus | .074 | Chronic renal failure | .107 | Disorder of lung | .091 | Female sexual arousal disorder | .076 | Diaphragmatic hernia | .044 | Osteoarthrosis involving multiple sites | .047 |
| Proteinuria | .064 | Arthralgia of the pelvic region and thigh | .092 | Impotence of organic origin | .086 | Chronic sinusitis | .061 | Degenerative joint disease of pelvis | .036 | Elevated levels of transaminase lactic acid dehydrogenase | .043 |
| Morbid obesity | .063 | Thoracic radiculitis | .071 | Pneumonia | .076 | Acute upper respiratory infection | .046 | Abnormal findings on diagnostic imaging of breast | .034 | Pure hyperglyceridemia | .034 |
| Benign essential hypertension | .038 | Joint pain | .036 | Overweight | .044 | Disease of liver | .043 | Goiter | .014 | Acrodermatitis continua | .021 |
| Vitamin D deficiency | .032 | Acute upper respiratory infection | .030 | Cholelithiasis without obstruction | .040 | Acute bronchitis | .004 | Benign essential hypertension | .002 | Diarrhea | .002 |
| Diabetic oculopathy associated with type 2 diabetes mellitus | .029 | Chronic rhinitis | .029 | Acute upper respiratory infection | .034 | Impacted cerumen | .002 | Conduction disorder of the heart | .001 | Peripheral venous insufficiency | .001 |
| CUMULATIVE PROBABILITY | .989 | | .998 | | .997 | | 1.000 | | 1.000 | | .992 |

As seen from the leftmost column in Table II, many of the diagnosed conditions grouped together in Topic A are clearly related to *Diabetes,* which is one of the most frequent causes of decrease in kidney function in the US [16]. It is well established medically that *Type 2 Diabetes* and *Hyperlipidemia* are closely associated conditions [17]. Similarly, *Type 2 Diabetes Mellitus*, *Benign Essential Hypertension* and *Morbid Obesity* are known to co-occur [18]. Moreover, *Vitamin D deficiency* is a common phenomenon in *Chronic Kidney disease* and hence frequently co-occurs with *Diabetes*. Likewise, most of the conditions grouped together in Topic B are related to *Limb* or *Joint pain,* conditions frequently occuring in patients suffering from advanced kidney disease, which explains the high probability of *Chronic Renal failure*, *Limb-* and *Joint-pain* to all be associated with the same topic [19]. We similarly assess the medical validity of each of the other topics identified by our model [16-19].

***Quantitative Evaluation:*** We next measure the quality of the resulting topics in terms of *tightness* and *distinctiveness*.

We assess the *tightness* of the topics by examining whether each can be specified by a small number of coded conditions. Thus for each topic we inspect the number of codes assigned a probability greater than a threshold value, set to 0.01. The observation that for each topic, only 10 or fewer of the 180 codes have a probability above 0.01, and that the cumulative probability of these 10 codes adds up to more than 0.9, indicates that the 10 conditions are sufficient for characterizing a topic, illustrating the tightness of the topics.

We assess the *distinctiveness* of the topics by calculating the inter-topic distance between all distinct pairs of 20 topics using the *JSD* [13] to measure how well-separated each topic is from another. The mean, median, and minimum values of the inter-topic distances obtained are 0.666, 0.692 and 0.483 respectively. As mentioned earlier, JSD values range between 0 and *ln(2)* (~0.693), where 0 indicates identical distributions, and *ln(2)* indicates non-overlapping distributions. The higher the JSD value between two topics, the more distinct they are from one another. The high mean and median values (close to the upper bound of *ln(2)*) of the inter-topic distances indicate that the majority of topic pairs obtained by our model are *distinct*.

## IV. CONCLUSION AND FUTURE WORK

We reported preliminary results obtained from our data-driven approach, using LDA to identify patterns of co-occurring medical conditions within an EMR dataset. Our results indicate that most of the coded conditions grouped together as topics are indeed known to co-occur according to the medical literature. We also quantitatively evaluate the performance of our method and demonstrate that the topics identified by our method are tight and distinct. Tightness is established by showing that each topic can be defined by ten or fewer conditions. Distinctiveness is established by illustrating that the large majority of the topic pairs are separated by a high Jensen-Shannon divergence.

We believe that our approach can be used to support clinical decision making. The data driven approach for identifying associated conditions can be used as a basis for a system that facilitates diagnosis and data entry in clinical settings by suggesting conditions that may co-occur with the patient's current diagnosed conditions.

ACKNOWLEDGMENT

This research was supported by the National Institute of General Medical Sciences, NIGMS IDeA grants U54-GM104941 and P20 GM103446. We thank James T. Laughery and Sarahfaye Heckler for their major role in building the dataset.